\documentclass[10pt]{article}
\usepackage[letterpaper, top=0.98in, bottom=0.98in, left=0.98in, right=0.98in]{geometry}
\usepackage{graphicx} 
\usepackage{layout}
\usepackage{blindtext}
\usepackage{multicol}
\usepackage{fancyhdr}
\usepackage{lastpage}
\usepackage{setspace}
\usepackage{ragged2e}
\usepackage{indentfirst}
\usepackage{titlesec}
\usepackage{mathptmx}
\usepackage{hyperref}
\usepackage{amsmath}
\usepackage{amssymb}
\usepackage{wrapfig}
\usepackage{caption}
\usepackage{multirow}
\usepackage{booktabs}
\usepackage{array}
\usepackage{tabularx}
\usepackage{svg}
\usepackage[numbers,sort&compress]{natbib}
\usepackage{arydshln}

\graphicspath{ {./diagrams/}{./diagrams/} }

\titlespacing\section{0pt}{12pt plus 4pt minus 2pt}{0pt plus 2pt minus 2pt}
\titlespacing\subsection{0pt}{12pt plus 4pt minus 2pt}{0pt plus 2pt minus 2pt}
\titlespacing\subsubsection{0pt}{12pt plus 4pt minus 2pt}{0pt plus 2pt minus 2pt}
\titlelabel{\normalsize\thetitle\quad}
\titleformat{\subsection}{\normalfont\normalsize\itshape}{\thesubsection}{1em}{}

\begin{document}
\onecolumn
\singlespacing
\pagestyle{fancy}

\fancyhf{}
\addtolength{\topmargin}{-5.15593pt}
\renewcommand{\headrulewidth}{0pt}
\fancyhead[C]{\fontsize{8}{9}\selectfont 74th International Astronautical Congress (IAC), Baku, Azerbaijan, 2-6 October 2023.\\Copyright ©2023 by Syn10. Published by the IAF, with permission and released to the IAF to publish in all forms.}
\fancyfoot[R]{Page \thepage\ of \pageref{LastPage}}
\fancyfoot[L]{IAC-23,B1,4,6,x76306}

\begin{Center}
IAC-23,B1,4,6,x76306
\end{Center}

\begin{Center}
\textbf{SatDM: Synthesizing Realistic Satellite Image with Semantic Layout Conditioning using Diffusion Models}
\end{Center}

\begin{Center}
\textbf{Orkhan Baghirli\textsuperscript{a}*, Hamid Askarov\textsuperscript{b}, Imran Ibrahimli\textsuperscript{c}, Ismat Bakhishov\textsuperscript{d}, Nabi Nabiyev\textsuperscript{e}}
\end{Center}

\begin{FlushLeft}
\textsuperscript{a} \emph{Syn10, Harju maakond, Tallinn, Kesklinna linnaosa, Ahtri tn 12, 15551}, \href{mailto:orkhan.baghirli@syn10.ai}{\nolinkurl{orkhan.baghirli@syn10.ai}}

\textsuperscript{b} \emph{Azercosmos, Space Agency of Republic of Azerbaijan, Uzeyir Hajibayli, 72}, \href{mailto:hamid.asgarov@azercosmos.az}{\nolinkurl{hamid.asgarov@azercosmos.az}}

\textsuperscript{c} \emph{Department of Informatics, University of Hamburg, HH 20148}, \href{mailto:imran.ibrahimli@studium.uni-hamburg.de}{\nolinkurl{imran.ibrahimli@studium.uni-hamburg.de}}

\textsuperscript{d} \emph{Azercosmos, Space Agency of Republic of Azerbaijan, Uzeyir Hajibayli, 72}, \href{mailto:ismatbakhishov@gmail.com}{\nolinkurl{ismatbakhishov@gmail.com}}

\textsuperscript{e} \emph{Department of Computer Science, Purdue University, West Lafayette, IN 47907}, \href{mailto:nnabiyev@purdue.edu}{\nolinkurl{nnabiyev@purdue.edu}}

\end{FlushLeft}

\section*{\centering\normalsize Abstract}
Deep learning models in the Earth Observation domain heavily rely on the availability of large-scale accurately labeled satellite imagery. However, obtaining and labeling satellite imagery is a resource-intensive endeavor. While generative models offer a promising solution to address data scarcity, their potential remains underexplored. Recently, Denoising Diffusion Probabilistic Models (DDPMs) have demonstrated significant promise in synthesizing realistic images from semantic layouts. In this paper, a conditional DDPM model capable of taking a semantic map and generating high-quality, diverse, and correspondingly accurate satellite images is implemented. Additionally, a comprehensive illustration of the optimization dynamics is provided. The proposed methodology integrates cutting-edge techniques such as variance learning, classifier-free guidance, and improved noise scheduling. The denoising network architecture is further complemented by the incorporation of adaptive normalization and self-attention mechanisms, enhancing the model's capabilities. The effectiveness of our proposed model is validated using a meticulously labeled dataset introduced within the context of this study. Validation encompasses both algorithmic methods such as Fréchet Inception Distance (FID) and Intersection over Union (IoU), as well as a human opinion study. Our findings indicate that the generated samples exhibit minimal deviation from real ones, opening doors for practical applications such as data augmentation. We look forward to further explorations of DDPMs in a wider variety of settings and data modalities. An open-source reference implementation of the algorithm and a link to the benchmarked dataset are provided at \href{https://github.com/obaghirli/syn10-diffusion}{https://github.com/obaghirli/syn10-diffusion}.

\noindent\textbf{Keywords:} generative models, conditional diffusion models, semantic image synthesis, building footprint dataset, remote sensing, satellite imagery
\vspace{1\baselineskip}

\begin{multicols}{2}
\section{\normalsize Introduction}\label{sec:intro}
Synthetic image generation is one of the fundamental challenges in the field of computer vision, with the goal of producing imagery that is indistinguishable from real images in terms of both fidelity and diversity. This task can also be seen as the inverse of semantic segmentation, where an input image is mapped to its corresponding semantic layout. Image generation can be categorized as unconditional or conditional. In the unconditional setting, the generative model relies solely on random noise as input, while in the conditional setting, additional information, such as semantic layouts, is provided. Conditional image generation has been extensively studied in the literature and has found many applications in various industries due to the well-defined nature of this paradigm and the better quality of the generated images compared to its unconditional counterpart.

Machine learning algorithms rely heavily on the availability of large-scale datasets to achieve optimal performance. The size of the dataset significantly impacts the discriminative and expressive power of these models. Most of the challenges encountered in the industry are formulated within a supervised training paradigm, where both the input data and the corresponding output labels are available to maximize performance. However, acquiring a large number of labeled samples is a laborious task that increases the development cost of the model. Moreover, manual labeling at scale is prone to random or systematic errors, which escalate with the dataset's scale and complexity. These errors can have detrimental effects on both the project budget and the model's performance.

In this paper, we propose a methodology for generating photorealistic synthetic imagery conditioned on semantic layout to augment existing datasets. We implement and deploy a novel denoising diffusion model on a dataset comprising optical satellite imagery and evaluate its performance both quantitatively and qualitatively. Optical satellite imagery features complex structural and textural characteristics, making it a more challenging domain compared to the commonly used benchmarks of natural images for generative modeling. Our results demonstrate that the denoising diffusion models can produce high-quality samples at multiple resolutions, even with limited data and computational resources. \end{multicols}\twocolumn

\noindent The primary contributions of this work include:
\begin{itemize}
\item The compilation of SAT25K, a meticulously curated building footprint dataset comprising image tiles and corresponding semantic layouts.
\item An in-depth exploration of diffusion models for generating synthetic satellite imagery.
\item The development of SatDM, a high-performance conditional diffusion model tailored for semantic layout conditioning. 
\item The release of source code and model weights to facilitate reproducibility.
\end{itemize}

The rest of this paper is organized as follows: In Section \ref{sec:review}, we review and summarize relevant existing literature to establish the groundwork for our research. Section \ref{sec:method} outlines the key components of our proposed methodology. The experimental setup is discussed in Section \ref{sec:exp}. Section \ref{sec:results} presents our findings, offering an in-depth discussion of the results. In Section \ref{sec:conclusion}, we examine the limitations of our work and outline potential directions for future research before concluding.

\section{\normalsize Related Work}\label{sec:review}
Generative Adversarial Networks (GANs) \cite{goodfellow2014generative} have been at the center of image generation for the past decade. GANs heavily benefited from the striking successes of the advancements in the field of deep learning, thus incorporating backpropagation and dropout algorithms on top of the multilayer perceptron architecture with piecewise linear units. The underlying premise of the GANs is simultaneously training the generator network with the discriminator network until the discriminator cannot distinguish between the real samples and samples generated by the generator network. GANs also did not suffer from the intractable probability density functions during the loss formulation as in the previous probabilistic models. Furthermore, sampling in GANs could be done seamlessly by using only forward propagation without involving approximate inference or Markov chains. Conditional GANs are introduced in \cite{Mirza2014ConditionalGA} by providing both the generator and discriminator networks with the conditioning signal via concatenation. A seminal work in conditional image generation known as pix2pix is proposed in \cite{Isola2016ImagetoImageTW} which allows the translation between different image domains. Following this work, \cite{Wang2017HighResolutionIS} synthesized high-resolution photorealistic images from semantic layouts through their more robust and optimized pix2pixHD architecture. For both of these architectures, conditioning information is fed to the generative and discriminate networks only once at the onset of the sequential convolutional neural networks (CNN) and normalization blocks. \cite{Miyato2018cGANsWP} takes a projection-based approach to condition the discriminator by computing the inner product between the embedded conditioning vector and feature vector, which significantly improves the quality of the conditional image generation. A revolutionary style-based GAN is introduced in \cite{Karras2018ASG} where the noise and embedded conditioning signal are injected into the synthesis network at multiple intermediate layers rather than the input layer only, as seen in many traditional generators. The style information in the form of a conditioning signal modulates the input feature map through the learned scale and bias parameters of the adaptive instance normalization (AdaIN) layers at multiple stages. A similar approach is undertaken in another highly influential work \cite{Park2019SemanticIS}, where the conditioning semantic layout is used to modulate the activations in normalization layers through spatially adaptive, learned transformation. This is in contrast to the traditional methods where the semantic layout is fed to the deep network as input to be processed through stacks of normalization layers, which tend to wash away the quality of the propagating signal. In a study to explore the scalability of GANs, authors in \cite{Brock2018LargeSG} successfully trained large-scale BigGAN-deep architectures and demonstrated that GANs benefit dramatically from scaling. Even though GANs have achieved significant success in generative modeling, their training is subject to many adversities. To this end, GANs capture less diversity and are often difficult to train, collapsing without carefully selected hyperparameters and regularizers \cite{Salimans2016ImprovedTF, Arjovsky2017WassersteinG, Karras2017ProgressiveGO, Kurach2018TheGL, Bau2018GANDV, Brock2018LargeSG, Dhariwal2021DiffusionMB}. Furthermore, objectively evaluating the implicit generative models such as GANs is difficult \cite{Theis2015ANO} due to their lack of tractable likelihood function. On the other hand, recent advancements have shown that likelihood-based diffusion models can produce high-quality images while offering desirable properties such as broader distribution coverage, a stationary training objective, and ease of scalability \cite{Dhariwal2021DiffusionMB}. 

In \cite{Welling2011BayesianLV}, a method for Bayesian learning from large-scale datasets is proposed.  This method involves the stochastic optimization of a likelihood through the use of Langevin dynamics. This process introduces noise into the parameter updates, leading the parameter trajectory to converge towards the complete posterior distribution, not just the maximum a posteriori mode. Initially, the algorithm resembles stochastic optimization, but it automatically shifts towards simulating samples from the posterior using Langevin dynamics. Unparalleled to Bayesian methods, in a groundbreaking work \cite{SohlDickstein2015DeepUL} inspired by non-equilibrium statistical physics, diffusion probabilistic models were introduced. These models enable the capturing of data distributions of arbitrary forms while allowing exact sampling through computational tractability. The core concept revolves around systematically and gradually disintegrating the patterns present in a data distribution using an iterative forward diffusion process. Subsequently, a reverse diffusion process that reconstructs the original structure in the data is learned. Learning within this framework involves estimating slight perturbations to the diffusion process. Following this work, a score-based generative modeling framework was proposed in \cite{Song2019GenerativeMB} consisting of two key components: score matching and annealed Langevin dynamics. The fundamental principle behind score-based modeling is perturbing the original data with varying levels of Gaussian noise to estimate the score, which represents the gradient of the log-density function at the input data point. A neural network is trained to predict this gradient field from the data. During sampling, annealed Langevin dynamics is used to progress from high to low noise levels until the samples become indistinguishable from the original data. A connection between score matching and diffusion probabilistic models is revealed in \cite{Ho2020DenoisingDP} as the authors show that under certain parameterization, denoising score matching models exhibit equivalence to diffusion probabilistic models. During learning the reverse diffusion process, the neural network parameterization, which predicts the noise levels in perturbed data rather than the forward diffusion process posterior mean, resulted in a superior sample quality. Thus, the injected noise parameterization leads to a simplified, weighted variational bound objective for diffusion models, resembling denoising score matching during training and Langevin dynamics during sampling. The authors concluded that despite the high sample quality of their method, the log-likelihoods are not competitive when compared to those of other likelihood-based models. Furthermore, since the diffusion process involves multiple forward steps to gradually destroy the signal, reversing the diffusion process to reconstruct the signal also necessitates numerous steps, resulting in a slow sampler. To address these difficulties around DDPMs, \cite{Nichol2021ImprovedDD} proposed several improvements. The authors suggested changing the linear noise scheduler to a cosine noise scheduler; thus maintaining a less abrupt diffusion process, learning model variance at each timestep rather than setting to a predetermined constant value, and switching the timestep sampler from uniform to importance sampler; thus improving log-likelihood, and adjusting the sampling variances based on the learned model variances for an arbitrary subsequence of the original sampling trajectory; thus improving the sampling speed. In a parallel work, rather than following the original DDPMs conceptualization, \cite{Song2020DenoisingDI} proposed a change to the underlying principles, which enables faster image generation. While retaining the original training objective of DDPMs, the authors redefined the forward process as non-Markovian, hence achieving much shorter generative Markov chains, leading to accelerated sampling with only a slight degradation in sample quality. In an effort to reduce the sampling time for diffusion models, \cite{salimans2022progressive} proposed a method resembling a distillation process, which is applied to the sampler of implicit models in a progressive way, halving the number of required sampling steps in each iteration. Authors in \cite{Dhariwal2021DiffusionMB} argue that one of the reasons why the diffusion models may still fall short of the quality of samples generated by GANs is that a trade-off mechanism between fidelity and diversity is incorporated into the GANs architecture, which allows them to produce more visually pleasing images at the cost of diversity, which is known as the truncation trick. To make DDPMs also benefit from this trade-off, they proposed auxiliary classifier guidance to the generation process inspired by the heavy use of class labels by conditional GANs and the role of estimated gradients from the data in the noise conditional score networks (NCSN). Following this study, \cite{Ho2021CascadedDM} demonstrated that high-resolution high-fidelity samples can be generated by cascading conditional diffusion-based models obviating the need for a companion classifier. Moreover, the proposed model exhibited superior performance compared to state-of-the-art GAN-based models. However, training multiple diffusion models and sampling sequentially is very time-consuming. In an attempt to achieve a GAN-like trade-off between the diversity and fidelity of the generated images without requiring an auxiliary classifier, authors in \cite{Ho2022ClassifierFreeDG} proposed a classifier-free guidance schema purely based on the diffusion models. In this schema, conditional and unconditional diffusion models are trained jointly without increasing the number of total training parameters, and sampling is performed using the linear combination of conditional and unconditional score estimates. Increasing the strength of this linear interpolation leads to an increase in sample fidelity and a decrease in sample diversity. To address the drawbacks of previous work, authors in \cite{Rombach2021HighResolutionIS} departed from working on the image space to compressed latent space of lower dimensionality through the pre-trained autoencoders, which made the high-resolution synthesis possible with significantly reduced computational requirements. The practicality of latent diffusion models (LDMs) opened the door to the development of large-scale diffusion models such as Stable Diffusion which is trained on billions of images conditioned on text prompts. To transfer the capabilities of general-purpose large diffusion models to more task-specific domains, where access to large amounts of training data is not feasible, authors in \cite{Zhang2023AddingCC} presented a control mechanism that allows the fine-tuning of the large models while preserving the knowledge extracted from billions of images.  

In the field of remote sensing, synthetic image generation has been receiving increasing attention. Researchers have explored various approaches to tackle the image generation task, and these approaches can be broadly categorized into three main groups: techniques employing GANs, DDPMs, and simulated sensors and environments. 

Annotated hyperspectral data generation has been addressed in \cite{Audebert2018GenerativeAN} using GANs, and the study validated the use of synthetic samples as an effective data augmentation strategy. River image synthesis for the purpose of hydrological studies also utilizes GANs to generate high-resolution artificial river images \cite{Gautam2020RealisticRI}. Previous work in regard to the generation of synthetic multispectral imagery \cite{Mohandoss2020GeneratingSM} and image style transfer from vegetation to desert \cite{Abady2020GANGO} using Sentinel-2 data suggested promising outcomes. With architectural and algorithmic modifications, authors reported satisfactory results and showed that GANs are capable of preserving relationships between different bands at varying resolutions. A novel self-attending task GAN is introduced in \cite{Toner2021SelfAttendingTG}, enabling the generation of realistic high-contrast scientific imagery of resident space objects while preserving localized semantic content.

Synthesizing images conditioned on additional information such as class labels, semantic layouts, and other data modalities grants more fine-tuned control over the generation process and allows us to ask the model to generate images of a specific type.  Translation from satellite images to maps \cite{Ganguli2019GeoGANAC, Ingale2021ImageTI}, and street views to satellite images \cite{Shah2021SatGANSI} utilizes the conditioning capability of GANs to generate the desired outcome. Another study \cite{Cardoso2022ConditionalPG} formulates the image generation task as the completion of missing pixels in an image conditioned on adjacent pixels. A study in \cite{Reyes2019SARtoOpticalIT} adopts GANs conditioned on reference optical images to enhance the interpretability of SAR data through SAR-to-optical domain translation. 

Another focus is on generating synthetic images using simulation platforms. Authors in \cite{Shermeyer2020RarePlanesSD} use a proprietary simulation software to generate overhead plane images with a novel placement on the map, and validate that detection models trained on synthetic data together with only a small portion of real data can potentially reach the performance of models trained solely on real data; thus reducing the need for annotated real data. 

The application of diffusion-based models to satellite imagery generation represents a relatively new and promising area of research in remote sensing. Motivated by the effective application of diffusion models in natural images and the extensive utilization of GANs for image super-resolution in remote sensing, researchers in \cite{Han2023EnhancingRS} adopted a diffusion-based hybrid model conditioned on the features of low-resolution image, extracted through a transformer network and CNN, to guide the image generation. Additionally, a recent diffusion-based model conditioned on SAR data for cloud removal task demonstrated promising results \cite{Zhao2023CloudRI}. In another study \cite{Cao2023DDRFDD}, the image fusion task is formulated as an image-to-image translation, where the diffusion model is conditioned on the low-resolution multi-spectral image, and high-resolution panchromatic image to guide the generation of a pansharpened high-resolution multi-spectral image. Authors in \cite{espinosa2023generate} demonstrate the conditioning of pre-trained diffusion models on cartographic data to generate realistic satellite images.

While diffusion models have shown promising potential and are gradually replacing traditional state-of-the-art methods in various domains, their application in conditioned image synthesis remains relatively underexplored, especially in the context of satellite imagery. This indicates a significant gap in the existing body of knowledge and presents an opportunity for further investigation.

\section{\normalsize Methodology}\label{sec:method}
Our goal is to design a conditional generative model that estimates the reverse of the forward diffusion process which converts complex data distribution into a simple noise distribution by gradually adding small isotropic Gaussian noise with a smooth variance schedule to the intermediate latent distributions throughout the sufficiently large diffusion steps.   
\subsection{\normalsize Loss Function}
To maintain consistency with the prior research and prevent potential confusion, we will omit the conditioning signal $c$ when deriving the loss function. The probability the generative model assigns to data in its tractable form can be evaluated as relative probability of the reverse trajectories $p(x_{t-1}|x_{t})$ and forward trajectories $q(x_{t}|x_{t-1})$, averaged over forward trajectories $q(x_{1:T}|x_{0})$ as in Eq. \ref{p:x_0}.
\begin{equation}\label{p:x_0}
\begin{split}
p_{\theta}(x_0) & = \int dx_{1:T} p_{\theta}(x_{0:T}) \\
 & = \int dx_{1:T} p_{\theta}(x_{0:T}) \frac{q(x_{1:T}|x_{0})}{q(x_{1:T}|x_{0})} \\
 & = \int dx_{1:T} q(x_{1:T}|x_{0}) p(x_T) \prod_{t=1}^{T} \frac{ p_{\theta}(x_{t-1}|x_{t}) }{q(x_{t}|x_{t-1})}
\end{split}
\end{equation}
where the reverse process is defined as a Markov chain with learned Gaussian transitions, starting at $p(x_{T}) = \mathcal{N} \left( x_{T}; \mathbf{0}, \mathbf{I} \right)$.

Training is performed by maximizing the evidence lower bound (ELBO) on log-likelihood $L$ (Eq. \ref{model:loglikelihood}), which can also be expressed as minimizing the upper bound on negative log-likelihood (Eq. \ref{model:negativeloglikelihood}).
\begin{align}\label{model:loglikelihood}
    L = \int dx_{0} q(x_{0}) \log( p_{\theta}(x_{0}) )
\end{align}
\begin{equation}\label{model:negativeloglikelihood}
\begin{split}
\mathbb{E}[&-\log(p_{\theta}(x_{0}))] \leq \mathbb{E}_{q}[-\log{\frac{p_{\theta}(x_{0:T})}{q(x_{1:T}|x_{0})}}]\\
    &= \mathbb{E}_{q}[- \log(p(x_{T})) - \sum_{t \geq 1} \log(\frac{p_{\theta}(x_{t-1}|x_{t})}{q(x_{t}|x_{t-1})})]
\end{split}
\end{equation}
\begin{equation}\label{model:terms}
    \mathbb{E}_{q}[L_{T} + \sum_{t>1} L_{t-1} + L_{0}]
\end{equation}
Rewriting the Eq. \ref{model:negativeloglikelihood} as Eq. \ref{model:terms} allows one to explore the individual contributions of the terms $L_{0}, L_{t-1}$, and $L_{T}$ (Eq. \ref{model: L0}, \ref{model: Ltminus1}, \ref{model: LT}) involved in the optimization process.
\begin{align}
    L_{0}   &= - \log(p_{\theta}(x_{0}|x_{1})) \label{model: L0} \\
    L_{t-1} &= D_{KL}(q(x_{t-1}|x_{t},x_{0}) || p_{\theta}(x_{t-1}|x_{t})) \label{model: Ltminus1} \\
    L_{T}   &= D_{KL}(q(x_{T}|x_{0}) || p(x_{T})) \label{model: LT}
\end{align}
Given sufficiently large diffusion steps $T$, infinitesimal and smooth noise variance $\beta_{t}$, forward noising process adequately destroys the data distribution so that $q(x_{T}|x_{0}) \approx \mathcal{N}(\mathbf{0}, \mathbf{I})$ and $p(x_{T}) \approx \mathcal{N}(\mathbf{0}, \mathbf{I})$; hence the $KL$ divergence between two nearly isotropic Gaussian distributions (Eq. \ref{model: LT}) becomes negligibly small. The $L_{0}$ term corresponds to the reverse process decoder, which is computed using the discretized cumulative Gaussian distribution function as described in \cite{Ho2020DenoisingDP}. The $L_{t-1}$ term, which dominates the optimization process, is composed of $KL$ divergence between two Gaussian distributions - the reverse process posterior $q(x_{t-1}|x_{t},x_{0})$ conditioned on data distribution $x_{0} \sim q(x_{0})$ and neural network estimate of reverse process $p_{\theta}(x_{t-1}|x_{t})$.
\begin{align}\label{beta}
\begin{split}
    \bar{\alpha}_{t} = \frac{f(t)}{f(0)}, \quad f(t) = \cos \left( \frac{t/T + s}{1+s} \cdot \frac{\pi}{2} \right)^2 \\
    \beta_{t} = 1-\frac{\bar{\alpha}_{t}}{\bar{\alpha}_{t-1}}, \quad \alpha_{t} = 1-\beta_{t}, \quad \bar{\alpha}_{t} = \prod_{s=0}^{t} \alpha_{s}
\end{split}
\end{align}
Reverse process posterior $q(x_{t-1}|x_{t},x_{0})$ can be derived using the Bayes theorem and Eqs. \ref{transitions_1} - \ref{transitions_3}. To prevent the abrupt disturbance in noise level, the cosine variance scheduler described in \cite{Nichol2021ImprovedDD} is adopted (Eq. \ref{beta}). Here, $s=0.008$ is a small offset to maintain numerical stability and $\beta_{t}$ is clipped to be between $0$ and $1$.   
\begin{align}
    q(x_{t}|x_{t-1}) &= \mathcal{N}(x_{t}; \sqrt{1-\beta_{t}}x_{t-1}, \beta_{t} \mathbf{I}) \label{transitions_1} \\
    q(x_{t}|x_{0}) &= \mathcal{N}(x_{t}; \sqrt{\bar{\alpha}_{t}}x_{0}, (1-\bar{\alpha}_{t}) \mathbf{I}) \label{transitions_2} \\
    q(x_{t-1}|x_{0}) &= \mathcal{N}(x_{t-1}; \sqrt{\bar{\alpha}_{t-1}}x_{0}, (1-\bar{\alpha}_{t-1}) \mathbf{I}) \label{transitions_3} 
\end{align}
Eq. \ref{posterior} shows that the reverse process posterior is parameterized by posterior mean $\widetilde{\mu}_{t}(x_{t}, x_{0})$ and posterior variance $\widetilde{\beta}_{t}$ described in Eq. \ref{posterior_mean} and Eq. \ref{posterior_variance}, respectively. 
\begin{align}
    q(x_{t-1}|x_{t}, x_{0}) &= \mathcal{N}(x_{t-1}; \widetilde{\mu}(x_{t}, x_{0}), \widetilde{\beta}_{t} \mathbf{I}) \label{posterior} \\
    \widetilde{\mu}_{t}(x_{t}, x_{0}) &= \frac{\sqrt{\bar{\alpha}_{t-1}} \beta_{t}}{1- \bar{\alpha}_{t}} x_{0} + \frac{ \sqrt{\alpha_{t}}(1-\bar{\alpha}_{t-1}) }{1- \bar{\alpha}_{t}} x_{t} \label{posterior_mean} \\
    \widetilde{\beta}_{t} &= \frac{1- \bar{\alpha}_{t-1}}{1- \bar{\alpha}_{t}} \beta_{t} \label{posterior_variance}
\end{align}
Since $q_(x_{t-1}|x_{t})$ is intractable, we are approximating it with a deep neural network surrogate Eq. \ref{reverse}. This representation of the estimate reverse process is parameterized by model mean $\mu_{\theta}(x_{t}, t)$ and model variance $\sigma_{t}^{2}$, described in Eq. \ref{reverse_mean} and \ref{reverse_variance}, respectively. The model mean is derived by following the noise parameterization $x_{t}(x_{0}, \epsilon) = \sqrt{\bar{\alpha}_{t}} x_{0} + \sqrt{1-\bar{\alpha}_{t}} \epsilon \quad \epsilon \sim \mathcal{N}(\mathbf{0}, \mathbf{I})$ and Eq. \ref{posterior_mean}. $\epsilon_{\theta}$ is a function estimator intended to predict $\epsilon$ from $x_{t}$. The model variance $\sigma_{t}^{2}$ is fixed to a predefined constant for each diffusion step and can take either of $\beta_{t}$ or $\widetilde{\beta}_{t}$, which are the upper and lower bounds on the variance, respectively.
\begin{align}
    p_{\theta}(x_{t-1}|x_{t}) &= \mathcal{N}(x_{t-1}; \mu_{\theta}(x_{t}, t), \sigma_{t}^{2} \mathbf{I}) \label{reverse} \\
    \mu_{\theta}(x_{t}, t) &= \frac{1}{\sqrt{\alpha_{t}}}\left( x_{t} - \frac{\beta_{t}}{\sqrt{1-\bar{\alpha}_{t}}}\epsilon_{\theta}(x_{t}, t) \right) \label{reverse_mean} \\
    \sigma_{t}^{2} &\in \{\beta_{t}, \widetilde{\beta}_{t}\} \label{reverse_variance}
\end{align}
The loss term we want to minimize is the $KL$ divergence between the reverse process posterior and estimate of the reverse process (Eq. \ref{model: Ltminus1}), which has a closed form solution leading to Eq. \ref{mean_loss}. Applying the noise parameterization to the posterior mean results in a simple training objective function $L_{simple}$ (Eq. \ref{L_simple}).
\begin{align}
    L_{t-1} &= \mathbb{E}_{q} \left[ \frac{1}{2 \sigma_{t}^{2}} \| \widetilde{\mu}_{t}(x_{t}, x_{0}) - \mu_{\theta}(x_{t}, t) \|^{2} \right] \label{mean_loss} \\
    L_{simple} &= \mathbb{E}_{t, x_{0}, \epsilon} \left[\| \epsilon - \epsilon_{\theta}(x_{t},t) \|^2 \right] \label{L_simple}
\end{align}
During the derivation of $L_{simple}$, model variance $\sigma_{t}^2$ was kept fixed meaning that it was not part of the learning process. However, \cite{Nichol2021ImprovedDD} suggests that even though in the limit of infinite diffusion steps, the choice of model variance would not affect the sample quality at all, in practice with a much shorter forward trajectory, learning the model variance $\Sigma_{\theta}(x_{t}, t)$ in Eq. \ref{variance_interpolation} can improve the model log-likelihood, leading to better sample quality. Due to the incorporation of the predicted variance term $v$ in the adjusted model of the reverse process (Eq. \ref{reverse_sigma_term}), where $v$ varies between 0 and 1 and serves as a linear interpolation factor between the upper and lower bounds on the log-variance, a new modified training objective function denoted as $L_{hybrid}$ is introduced in Eq. \ref{L_hybrid}.
\begin{align}
    \Sigma_{\theta}(x_{t}, t) &= \exp\left(v \log(\beta_{t}) + (1-v) \log(\widetilde{\beta}_{t})\right) \label{variance_interpolation} \\
    p_{\theta}(x_{t-1}|x_{t}) &= \mathcal{N}(x_{t-1}; \mu_{\theta}(x_{t}, t), \Sigma_{\theta}(x_{t}, t)) \label{reverse_sigma_term} 
\end{align}
While $L_{simple}$ exclusively guides the update of the mean $\mu_{\theta}(x_{t}, t)$ assuming a fixed variance schedule (Eq. \ref{reverse}), $L_{vlb}$ exclusively facilitates the update of the variance $\Sigma_{\theta}(x_{t}, t)$, keeping the mean unaffected.  Optimizing $L_{vlb}$ is more challenging than optimizing $L_{simple}$ due to increased gradient noise. To address this, the hybrid loss function $L_{hybrid}$ combines these two approaches using a small weight parameter $\lambda = 0.001$. This combination allows for learning the variance while maintaining the overall stability of the optimization process.
\begin{equation}\label{L_hybrid}
    \begin{split}
        L_{vlb} &= L_{0} + L_{1} + \ldots + L_{T-1} + L_{T} \\
        L_{hybrid} &= L_{simple} + \lambda L_{vlb} 
    \end{split}
\end{equation}
We adopted the classifier-free guidance strategy proposed in \cite{Ho2022ClassifierFreeDG}, which jointly trains the unconditional model $\epsilon_{\theta}(x_{t}, t, c=\emptyset)$ and conditional model $\epsilon_{\theta}(x_{t}, t, c)$, simply by setting the conditioning signal $c$ to the unconditional class identifier $\emptyset$ with some probability $p_{uncond} = 0.2$, set as a hyperparameter. 
\begin{figure*}[t!]
    \centering
    \includegraphics[width=0.7\textwidth]{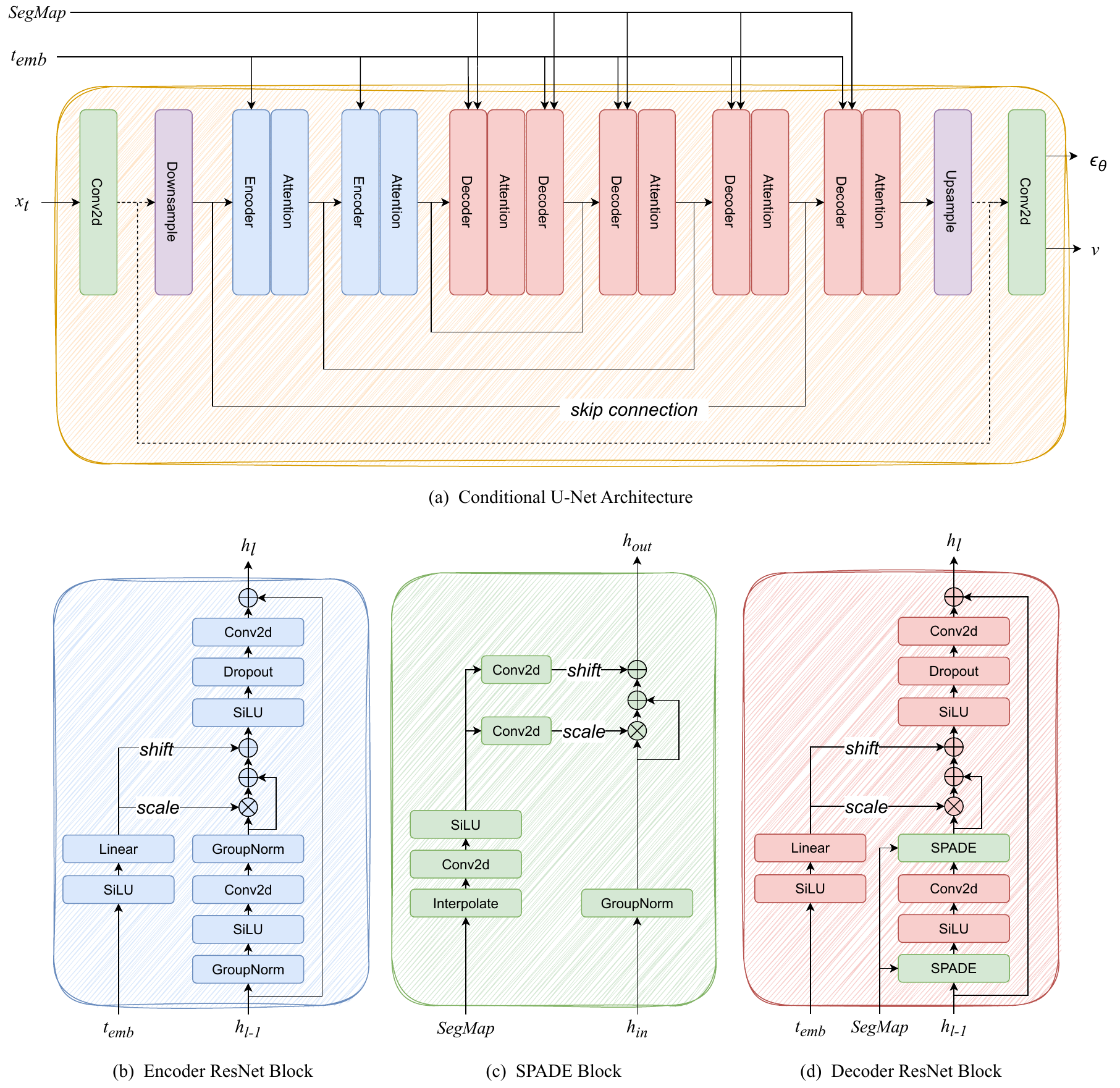}
    \caption{Denoising Network. (a) Conditional U-Net model estimates the noise mean and variance of a noisy input image $x_{t}$ at timestep t. (b) Encoder ResNet block captures the low-level representation of $x_{t}$. (c) SPADE block effectively modulates the input signal based on the conditioning SegMap, which is denoted as $c$ within the equations. (d) Decoder ResNet block incorporates the latent representation from the encoder block along with the semantic information to reconstruct the noise model.}
    \label{fig:model}
\end{figure*}
\subsection{\normalsize Sampling}
The sampling procedure starts at timestep $T$ with $x_{T} \sim \mathcal{N}(\mathbf{0}, \mathbf{I})$ drawn from a standard normal prior, which is then passed as input to the denoising neural network model $\widetilde{\epsilon}_{\theta}(x_{t}, t, c)$ as described in Eq. \ref{updated_epsilon}. 
\begin{align}\label{updated_epsilon}
\begin{split}
    \widetilde{\epsilon}_{\theta}(x_{t}, t, c) &= \epsilon_{\theta}(x_{t}, t, c=\emptyset)\\
    & + \omega \left(\epsilon_{\theta}(x_{t}, t, c) - \epsilon_{\theta}(x_{t}, t, c=\emptyset) \right)
\end{split}
\end{align}
which is a linear combination of conditional and unconditional noise estimates. The interpolation factor $\omega$ enables the trade-off between sample fidelity and diversity. 

The sampling process is conducted iteratively, where the output from the denoising network constitutes the input to the model at the subsequent iteration. To sample $x_{t-1} \sim p_{\theta}(x_{t-1}|x_{t}, c)$ is to compute Eq. \ref{sample_xtminus1}, where the noise profile $z$ follows Eq. \ref{sample_z}.
\begin{align}\label{sample_xtminus1}
\begin{split}
    x_{t-1} &= \frac{1}{\sqrt{\alpha_{t}}}\left( x_{t} - \frac{1-\alpha_{t}}{\sqrt{1-\bar{\alpha}_{t}}} \widetilde{\epsilon}_{\theta}(x_{t}, t, c) \right)\\
    & + \sqrt{\Sigma_{\theta}(x_{t}, t, c)} z
\end{split}
\end{align}
\begin{align}\label{sample_z}
    z = \begin{cases} 
      \mathcal{N}(\mathbf{0},\mathbf{I}) & t > 1 \\
      0 & t \leq 1 
    \end{cases}
\end{align}
 At the end of the sampling process, model mean $\mu_{\theta}(x_{1}, 1)$ is displayed noiselessly. 

\subsection{\normalsize Denoising Network}
Our model architecture is based on previous work \cite{Wang2022SemanticIS}, and builds upon the underlying time-conditional U-Net architecture, which is illustrated in Fig. \ref{fig:model} (a). The encoding path of the model, which consists of consecutive residual networks (ResNet) at each level (Fig. \ref{fig:model} (b)), captures the low-level latent representation of the noisy image $x_{t}$ at diffusion step $t$. The decoding path of the network, which consists of modified ResNet blocks (Fig. \ref{fig:model} (d)) that incorporate the conditioning segmentation map (SegMap) through the spatially-adaptive (SPADE) normalization module (Fig. \ref{fig:model} (c)), reconstructs the level of noise added to the original image $x_{0}$. In this manner, the semantic information encoded in SegMap is preserved, rather than being washed away after successive passes through the convolutional and normalization layers \cite{Karras2018ASG, Park2019SemanticIS}. Group normalization is applied as a normalization technique to reduce sensitivity to variations in batch size. The timestep embedding $t_{emb}$ modulates the input signal that is fed into both the encoder and decoder ResNet blocks, whereas the semantic information SegMap is exclusively injected into the network through the decoder ResNet blocks \cite{Zhang2023AddingCC}. Multi-head self-attention modules are added on top of the ResNet blocks \cite{Wang2017NonlocalNN} only at predefined resolutions such as 32, 16, and 8. The upsampling block first scales up the input using the nearest neighbors method, and then it performs a convolution operation. The downsampling block employs the convolution operation with a stride parameter set to 2. The decoding and encoding paths are interlinked via the skip connections to fuse information from both ends. The proposed conditional U-Net model estimates the noise mean $\epsilon_{\theta}(\cdot, t)$ (Eq. \ref{L_simple}) and $v$ component (Eq. \ref{variance_interpolation}), which is normalized to a range of 0 to 1 and interpolates between the upper and lower bounds on log-variance. 

\begin{table}[t]
\caption{Description of SAT25K dataset.}
\centering
\begin{tabular}{l c}
\toprule
Data Source & Google Earth \\
Satellite & Pleiades \\ 
Correction Method & Orthophotomosaic \\
Region & Mardakan, Baku, Azerbaijan \\
Coordinate System & EPSG:32639 - WGS 84 \\
Bands & Red, Green, Blue \\
Data Type & uint8 \\
Data Range & 0 - 255 \\
Resolution [m] & 0.5 \\
Annotation Classes & Binary \\
Number of Polygons & 25000 \\
Surface Area [$km^{2}$] & 92.3 \\
Footprints Area [$km^{2}$] & 3.98 \\
Pixel Count [M] & 370 \\
\bottomrule
\end{tabular}
\label{table:raw_data_stat}
\end{table}

\section{\normalsize Experiments}\label{sec:exp}
\subsection{\normalsize Dataset}
Due to the unavailability of high-precision annotated satellite imagery of building footprints that can serve as a benchmark dataset at the time of conducting the experiments, we opted to create our own. It is worth mentioning that manually labeling high-resolution imagery is an expensive and labor-intensive endeavor. In this section, we introduce the SAT25k dataset, which is composed of the high-resolution satellite imagery of buildings and the corresponding building footprint annotations (Table \ref{table:raw_data_stat}). 

\begin{table}[ht]
\caption{Distribution of the Organic and Augmented image tiles across the train and test splits of SAT25K dataset.}
\centering
\begin{tabular}{@{\extracolsep{4pt}}l c c c c@{}}
\toprule
\multirow{2}{*}{Split} & \multicolumn{2}{ c }{64 $\times$ 64} & \multicolumn{2}{ c }{128 $\times$ 128}  \\ 
\cline{2-3} \cline{4-5}
 & Org. & Aug. & Org. & Aug. \\
\midrule
Train & 24366 & 25634 & 24681 & 25319 \\ 
Test  & 5000 & 0 & 5000 & 0 \\ 
\bottomrule
\end{tabular}
\label{table:org_aug}
\end{table}
The dataset covers an area primarily characterized by suburban houses, complemented by a subset of industrial facilities and buildings, although not exhaustive in its representation. The designated region falls within the parameters of a semi-arid climatic zone. The features of interest remain unaffected by meteorological elements such as snow coverage, cloud presence, or any other adverse atmospheric conditions. The associated annotations are encoded in a pixel-wise binary format, where the values 1 and 0 correspond to the positive and negative classes, respectively.

The original dataset is tiled into $128 \times 128$ windows with $50\%$ overlap between the adjacent tiles. Subsequently, we downsampled the $128 \times 128$ tiles using the pixel area interpolation method to produce the $64 \times 64$ tiles. Tiles with less than $1\%$ positive class ratio are dropped. Both datasets at the resolution of $128 \times 128$ and $64 \times 64$ undergo a process of augmentation by applying the basic geometric transformations including random rotation $90^{\circ}$ and vertical flip, only on their training splits. The composition of the resulting datasets for both experiments is described in Table \ref{table:org_aug}.

\subsection{\normalsize Environment}
The experimentation was performed using a single NVIDIA Tesla V100 instance for the $128 \times 128$ resolution and a single NVIDIA Quadro P5000 instance for the $64 \times 64$ resolution. Both instances are equipped with 16GB of GPU memory. The remaining environment parameters are described in Table \ref{table:setup}.

\begin{table}[ht]
\caption{Environment parameters.}
\centering
\begin{tabularx}{0.93\columnwidth}{ 
   >{\raggedright\arraybackslash}X 
   >{\centering\arraybackslash}X 
   >{\centering\arraybackslash}X }
\toprule
  Parameter  & 64 $\times$ 64 & 128 $\times$ 128 \\
\midrule
    OS & Ubuntu 18.04 & Ubuntu 20.04 \\
    GPU model & Quadro P5000 & Tesla V100 \\
    GPU memory & 16GB & 16GB \\
    GPU count & 1 & 1 \\
    CPU model & Core I9 9900K & Xeon E5-2690 \\
    CPU memory & 128GB & 110GB \\
\bottomrule
\end{tabularx}
\label{table:setup}
\end{table}

\subsection{\normalsize Training}
The experiments at different resolutions are conducted on separate environments with parameters described in Table \ref{table:train_params}. The models are trained for 92 hours and 163 hours for experiments at resolutions of $64 \times 64$ and $128 \times 128$, respectively. For both experiments, we utilize the AdamW implementation of PyTorch 2.0 as the optimizer, with the default parameters except for the weight decay, which is configured at 0.05. Before the parameter update, the gradient norm is clipped to a maximum threshold of 1 to prevent unstable optimization. Furthermore, to ensure parameter updates remain robust against sudden fluctuations, we calculate exponential moving averages (EMA) of the model parameters, using a decay parameter set to 0.9999. For the early stages of the optimization is characterized as noisy, we added a delay of 10K iterations to start gathering the EMA values. During experimentation, we observed that the cosine annealing with a warm restarts scheme as a learning rate strategy hurts the optimization process at the inflection points, leading to instabilities. Therefore, the learning rate is configured such that it follows the cosine curve starting at the initial value of $2 \times 10^{-5}$, and decreasing down to 0 throughout the entire optimization period, without any restarts. The input tiles are scaled to a range of -1 to 1, while the SegMap is one-hot encoded. For each batch, the diffusion timesteps are sampled uniformly. The GPU utilizations are recorded at $93\%$  and $99\%$ for experiments at resolutions of $128 \times 128$ and $64 \times 64$, respectively.

\begin{table}[ht]
\caption{Training parameters.}
\centering
\begin{tabular}{l c c}
\toprule
  Parameter  & 64 $\times$ 64 & 128 $\times$ 128 \\
\midrule
    Diffusion Steps & 1000 & 1000 \\
    Noise Schedule & cosine & cosine \\
    Model Size & 31M & 130M \\
    Model Channels & 64 & 128 \\
    Depth & 3 & 4 \\
    Channel Multiplier & 1, 2, 3, 4 & 1, 1, 2, 3, 4 \\
    Head Channels & 32 & 64 \\
    Attention Resolution & 32, 16, 8 & 32, 16, 8 \\
    Number of ResNet blocks & 2 & 2 \\
    Dropout & 0.1 & 0 \\
    Batch Size & 32 & 8 \\
    Iterations & 400K & 1250K \\
    Initial Learning Rate & $2 \times 10^{-5}$ & $2 \times 10^{-5}$ \\
\bottomrule
\end{tabular}
\label{table:train_params}
\end{table}

\begin{figure*}[ht]
    \centering
    \includegraphics[width=\textwidth]{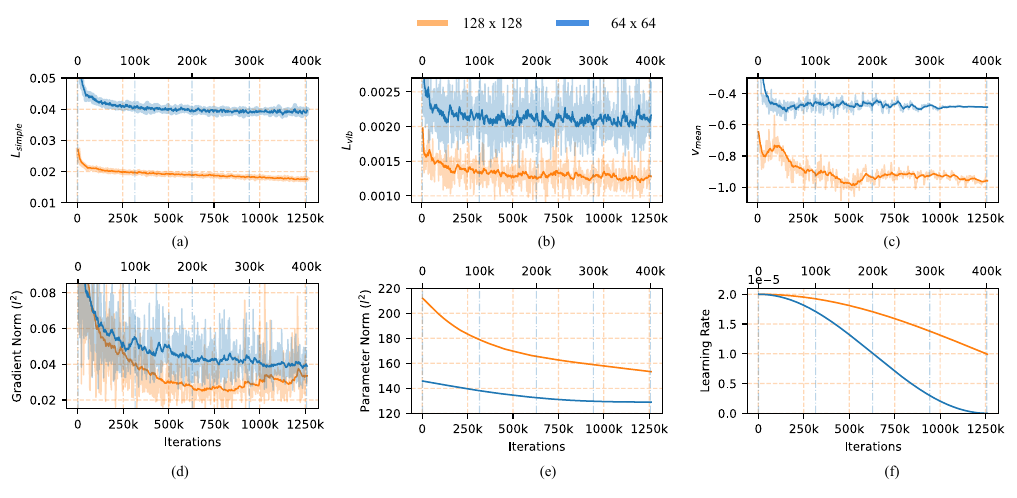}
    \caption{Optimization dynamics. (a) $L_{simple}$ term is computed as the mean squared error between the actual and estimated noise terms (Eq. \ref{L_simple}). (b) $L_{vlb}$ is calculated as the negative log-likelihood loss for term $L_{0}$ and KL divergence loss for terms $L_{1:T}$ (Eq. \ref{L_hybrid}). (c) The mean of the interpolation variable $v$, obtained directly from the model output, serves as the basis for estimating variance. (d) Monitoring of the gradient norms to identify anomalies such as exploding or vanishing gradients. (e) The parameter norm is calculated as the $L^{2}$-norm of all trainable model parameters. (f) Learning rate profile under the cosine annealing without warm restart schedule.}
    \label{fig:tensorboard}
\end{figure*}

The goal of optimization is to minimize the objective function $L_{hybrid}$ described in Eq. \ref{L_hybrid}. Fig. \ref{fig:tensorboard} illustrates the evolution of key indicators throughout the optimization process. Fig. \ref{fig:tensorboard} (a) shows that $L_{simple}$ follows a stable downward trajectory for both experiments, whereas model 64 demonstrates elevated losses. Additionally, despite model 64 being trained with larger batch size, its $L_{simple}$ curve exhibits more noise compared to that of model 128. It can be inferred that downsampling from $128 \times 128$ to $64 \times 64$ during the dataset preparation resulted in an irreversible loss of information, suggesting that higher resolutions imply a smaller and smoother loss term. The variational loss term $L_{vlb}$, which is portrayed in Fig. \ref{fig:tensorboard} (b), displays more turbulent convergence dynamics compared to $L_{simple}$, confirming the findings from an earlier study \cite{Nichol2021ImprovedDD}. The average of variance signal $v$, which is driven by the $L_{vlb}$ term and described in Eq. \ref{variance_interpolation}, is depicted in Fig. \ref{fig:tensorboard} (c) in its unnormalized form. Even though the $v$ term is not explicitly constrained, it follows a well-defined behavior. Both models follow the lower bound on the variance schedule $\widetilde{\beta}_{t}$, with model 128 exhibiting a closer alignment. Fig. \ref{fig:tensorboard} (d) demonstrates that as the optimization progresses, the gradient norms tend to decrease. Smaller gradient norms indicate that the changes in the parameter values are becoming smaller, implying that the algorithm is getting closer to an optimal solution. Over the course of training, the optimizer gradually adjusts the model parameters towards zero, which is reflected in Fig. \ref{fig:tensorboard} (e). The cosine annealing schedule (Fig. \ref{fig:tensorboard} (f)) reduces sensitivity to the choice of the initial learning rate and ensures a smooth transition to promote stable convergence.

\subsection{\normalsize Sampling}
Sampling is performed according to the parameters described in Table \ref{table:sample_params}. EMA of the model parameters are employed instead of their instantaneous snapshots during the sampling process. Since we are traversing every diffusion step during sampling, generating a dataset of 5000 synthetic images at resolutions of $64 \times 64$ and $128 \times 128$ takes substantial time. The average wall-clock time measured to generate each sample increases in accordance with the model size and length of the reverse process. The guidance scale $\omega$ is set to 1.5 for a balanced trade-off between sample fidelity and diversity \cite{Wang2022SemanticIS}. The GPU utilization for both models is recorded at $99\%$ across the entire sampling process. 

\begin{table}[ht]
\caption{Sampling parameters.}
\centering
\begin{tabular}{l c c}
\toprule
  Parameter  & 64 $\times$ 64 & 128 $\times$ 128 \\
\midrule
    GPU model & Quadro P5000 & Tesla V100 \\
    GPU count & 1 & 1 \\
    GPU utilization & $99\%$ & $99\%$ \\
    Batch Size & 256 & 64 \\
    Sampling Steps & 1000 & 1000 \\
    Guidance ($\omega$) & 1.5 & 1.5 \\
    Number of Samples & 5000 & 5000 \\
    Duration [hours] & 24.3 & 42.5 \\
    Throughput [$s^{-1}$] & $5.72 \times 10^{-2}$ & $3.26 \times 10^{-2}$ \\
\bottomrule
\end{tabular}
\label{table:sample_params}
\end{table}

\section{\normalsize Results}\label{sec:results}
\begin{figure*}[ht]
    \centering
    \includegraphics[width=\textwidth]{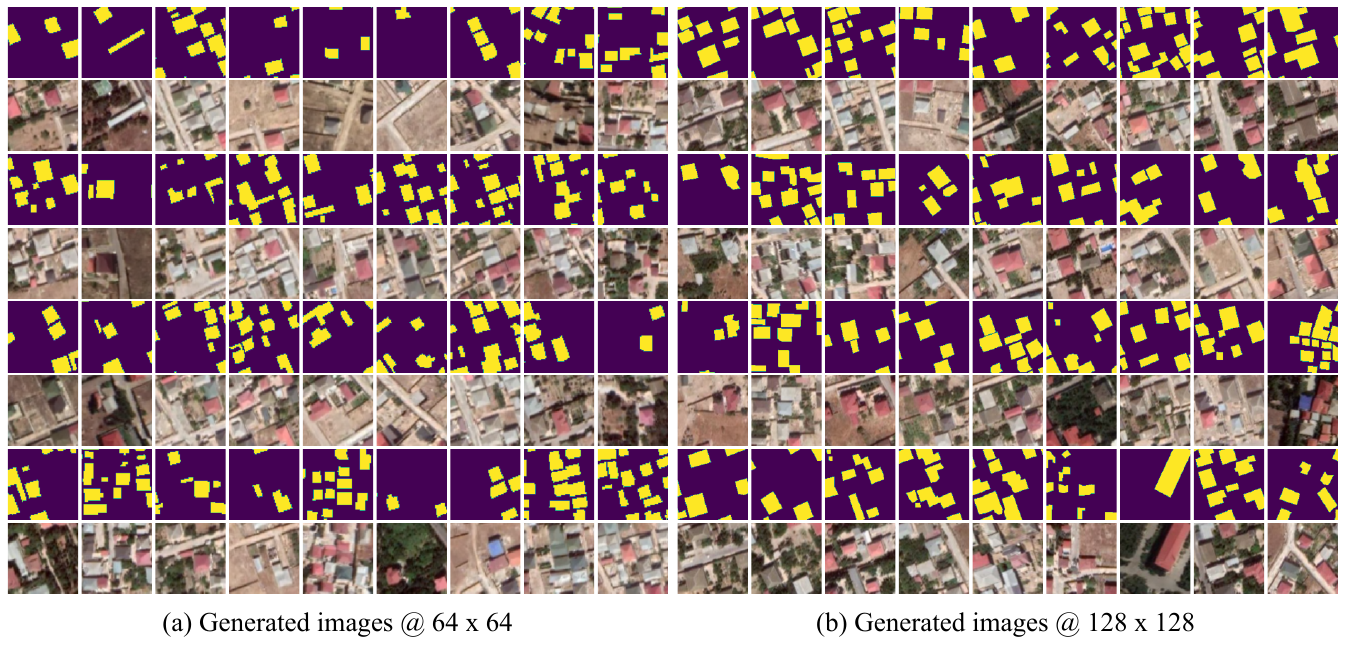}
    \caption{Model generated samples. The alternating rows represent the conditioning segmentation maps and generated images, respectively. (a) Samples generated at $64 \times 64$ resolution. (b) Samples generated at $128 \times 128$ resolution.}
    \label{fig:samples}
\end{figure*}

The generated images conditioned on their corresponding segmentation maps are illustrated in Fig. \ref{fig:samples}. The results indicate that both models are capable of synthesizing semantically meaningful, diverse, and high-fidelity samples. Furthermore, the strong alignment observed between the generated images and semantic maps signifies the successful integration of the conditioning mechanism into the model architecture. The proposed diffusion models efficiently capture the structure, style, texture, and color composition of the real images without displaying visual cues of overfitting or mode collapse. Notably, the models effectively handle the challenges inherent to satellite imagery, including object occlusion, shadows, straight lines, and the complex interdependence between spectral bands.

\begin{figure*}[ht]
    \centering
    \includegraphics[width=0.6\textwidth]{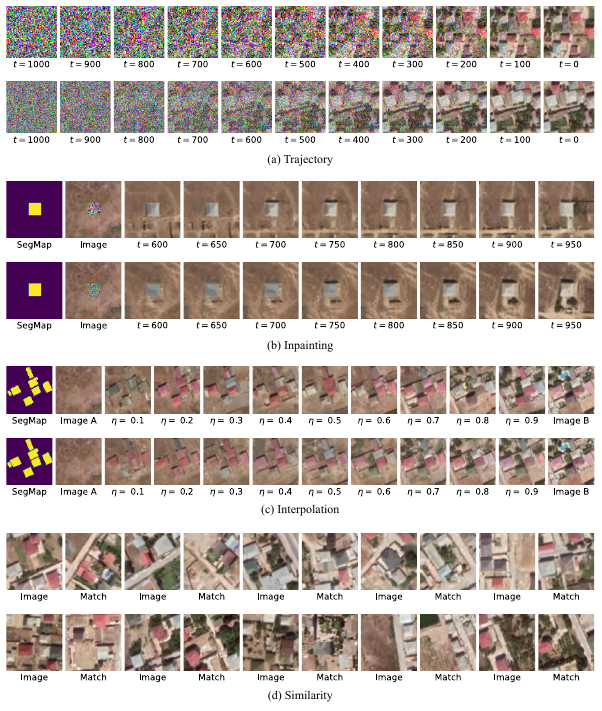}
    \caption{Analysis techniques. The top row corresponds to the $64 \times 64$ experiment, while the bottom row represents the $128 \times 128$ experiment. (a) An evolution of the image generation process over a reverse diffusion trajectory represented at various diffusion timesteps. (b) Inpainting of the partially corrupted image conditioned on a semantic map. (c) Interpolation between Image A and Image B, given the SegMap corresponding to Image B, where $\eta$ denotes the interpolation strength. (d) Similarity search between the generated and real samples. The generated image and its closest match from the training dataset are portrayed side-by-side.}
    \label{fig:results}
\end{figure*}

\subsection{\normalsize Fréchet inception distance}
 FID is a metric that provides a balanced assessment of image fidelity and diversity simultaneously. The FID scores described in Table \ref{table:perf_metrics} are measured between the 5000 synthetic samples generated by our proposed SatDM models and the 5000 real test images from the SAT25K dataset. The test images and their corresponding semantic labels have never participated in the training phase. The Inception V3 model trained on the ImageNet 1K dataset is used to calculate the FID scores. The $64 \times 64$ model outperformed the $128 \times 128$ model since the latter poses a more challenging optimization problem due to the increased dimensionality. The results of both resolutions are on par with the state-of-the-art. 

\begin{table}[ht]
\caption{Performance Metrics. The synthetic dataset represents the 5000 model-generated samples. SAT25K represents the test split of the corresponding dataset consisting of 5000 samples. The FID score is measured between the Synthetic and SAT25K datasets, while the IoU score is computed for each dataset separately. $\uparrow$ indicates the higher the better, while $\downarrow$ indicates the lower the better.}
\centering
\begin{tabular}{l c c c}
\toprule
 Dataset & Resolution & mIoU $\uparrow$ & FID $\downarrow$ \\
\hline \noalign{\vskip 1mm} 
    Synthetic & $64 \times 64$ & 0.48 & \multirow{2}{*}{25.6} \\
    SAT25K & $64 \times 64$ & 0.37 & \\ 
    \noalign{\vskip 1mm} 
    \hline
    \noalign{\vskip 1mm} 
    Synthetic & $128 \times 128$ & 0.60 & \multirow{2}{*}{29.7} \\
    SAT25K & $128 \times 128$ & 0.46 & \\
\bottomrule
\end{tabular}
\label{table:perf_metrics}
\end{table}

\subsection{\normalsize Intersection over Union}
The IoU quantifies the degree of overlap between the predicted segmentation map and the ground truth segmentation map. In the generation of synthetic samples, our diffusion model accepts pure noise and the ground truth segmentation map (SegMap) as input parameters. Subsequently, the generated sample is passed through an off-the-shelf segmentation model known as the Segment Anything Model (SAM), without undergoing any fine-tuning. The IoU scores, as described in Table \ref{table:perf_metrics}, are computed by comparing the predicted segmentation map of the generated images (SamMap-Synthetic) to the true mask (SegMap). The SAM model is configured to use 5 points for each connected component in the SegMap. To assess the fitness of the SAM model to the domain of satellite imagery from the test split of the SAT25K dataset, we also calculated the IoU score between the SAM`s semantic segmentation of the real imagery (SamMap-Real) and the corresponding SegMap. The results indicate that SAM, without any fine-tuning, can only offer limited performance in this context. Based on our findings, we observed that the $128 \times 128$ model outperformed the $64 \times 64$ model, primarily due to the degraded boundary structure of objects in the downsampled $64 \times 64$ dataset. 

\begin{table}[ht]
\caption{Human Opinion Study. The results are calculated individually for each respondent and then averaged across all respondents. Both questionnaires feature a balanced distribution of samples, with an equal representation of AI-generated and real images. A score of 0.5 represents maximum uncertainty in the context of binary classification.}
\centering
\begin{tabular}{l c c}
\toprule
   Metrics & 64 $\times$ 64 & 128 $\times$ 128 \\
\hline
    Accuracy & 0.57 & 0.53 \\
    Precision & 0.57 & 0.53 \\
    Recall & 0.60 & 0.50 \\
    F1 & 0.58 & 0.50 \\ 
\bottomrule
\end{tabular}
\label{table:human}
\end{table}

\subsection{\normalsize Human Opinion Study}
While the FID score is widely used to assess the performance of the generative models, it is limited in its ability to accurately evaluate various aspects of generated images, such as color accuracy, textual quality, semantic alignment, presence of artifacts, and subtle details. To address these limitations, we conducted a human opinion study, where we tasked 12 trained Geographic Information Systems (GIS) professionals to discriminate between the AI-generated and real samples. 

In this study, we designed two separate two-alternative forced choices (2AFC) questionnaires, one for each set of images at the resolutions of $64 \times 64$ and $128 \times 128$, featuring the participation of 9 out of 12 respondents and 10 out of 12 respondents, respectively. Each questionnaire included 25 randomly selected real samples from the test split of the SAT25K dataset and 25 model-generated samples. To enhance visual perceptibility, we resized the $64 \times 64$ samples to $128 \times 128$ using the nearest neighbors method. Respondents were asked to determine whether a given image was real or AI-generated, with a standardized time limit of 5 seconds for each question.

The optimum theoretical outcome of this study would be when the respondents are absolutely indecisive in their responses, to the extent of making completely random guesses. The results of the study are summarized in Table \ref{table:human}, where any deviation from 0.5, which corresponds to random guessing, indicates that humans were able to discern specific discriminative features between AI-generated and real samples. The results suggest that the samples at the resolution of $128 \times 128$ are more effective at deceiving humans.

\subsection{\normalsize Trajectory}
Fig. \ref{fig:results} (a) illustrates the reverse trajectory of image synthesis, starting as a pure isotropic Gaussian noise at timestep $t=1000$ and gradually evolving into a noise-free image at timestep $t=0$ through subsequent denoising operations. The results suggest that, in the early phases of the generation cycle, the $128 \times 128$ model learns the image's underlying structure, texture, and color composition more quickly compared to its $64 \times 64$ counterpart. Consequently, the $128 \times 128$ model dedicates a larger portion of its capacity to resolving high-frequency details present in the images, while the $64 \times 64$ model focuses more on resolving low-frequency details, given the relatively reduced presence of fine-grained information at this resolution.

\subsection{\normalsize Inpainting}
Fig. \ref{fig:results} (b) demonstrates the inpainting process, where the image is cut according to the conditioning segmentation map, and the hole is filled with noise. The partially corrupted image undergoes a forward diffusion process at various timesteps. The reconstructed images reveal that both models have developed a deep understanding of object and scene semantics. Images generated over longer trajectories exhibit more detailed object restoration while causing greater modifications to the surrounding scene. In contrast, shorter trajectories produce less detailed object restoration with minimal adjustments to the surrounding area.

\subsection{\normalsize Interpolation}
Interpolation between two images is depicted in Fig. \ref{fig:results} (c). Image A presents a minimalistic scene characterized by bare land cover and the absence of foreground objects, while Image B represents a complex scene. Both images undergo the diffusion process up to the predefined timestep $t=600$. The linear combination of the corrupted images, modeled as $x_{t} = x^{(A)}_{t} (1-\eta) + x^{(B)}_{t} \eta$, where $\eta$ denotes the interpolation strength, is fed to the denoising model along with the conditioning SeqMap of Image B. The results demonstrate that as $\eta$ increases, the generated images increasingly incorporate content and style from Image B. The semantically meaningful transitions from Image A to Image B signify the continuity of the denoising networks` latent space as the traversal of the latent space is reflected in smooth transitions in the data space.  

\subsection{\normalsize Similarity}
The generated images are assessed for potential overfitting by comparing them to the entire training split of the SAT25K dataset. For this purpose, a generated sample is encoded using the pre-trained Inception V3 model, and the resulting embedding vector is compared to the embeddings of real samples, with cosine similarity employed as the comparison metric. The closest matching images are showcased in Fig. \ref{fig:results} (d). The results reveal that both models generate distinct samples without exhibiting visual indicators of overfitting. 

\section{\normalsize Conclusions}\label{sec:conclusion}
In this paper, we introduce SatDM, a diffusion-based generative model that is conditioned on semantic layout. In addition, within the scope of this study, we introduce SAT25K, a novel building footprint dataset. Our models achieve state-of-the-art performance in the synthesis of realistic satellite imagery, excelling in both quantitative and qualitative assessments, paving the way for several intriguing applications. The proposed sampler is computationally intensive, and we leave the synthesis at higher resolutions for future investigations.

\section*{\normalsize Acknowledgements}
We are grateful for the generous support and sponsorship provided by Microsoft for Startups Founders Hub. This support has significantly contributed to the success of our research and development efforts. 

\renewcommand{\refname}{\normalsize References}
\bibliography{references}

\end{document}